\documentclass[conference]{IEEEtran}
\IEEEoverridecommandlockouts
\usepackage{cite}
\usepackage{amsmath,amssymb,amsfonts}
\usepackage{algorithmic}
\usepackage{graphicx}
\usepackage{textcomp}
\usepackage{xcolor}
\usepackage[hidelinks]{hyperref}
\usepackage{placeins}
\usepackage[caption=false,font=footnotesize]{subfig}
\usepackage{float}
\usepackage{caption}
\usepackage[font=footnotesize]{caption}
\usepackage{booktabs} 
\usepackage{wrapfig}

\def\BibTeX{{\rm B\kern-.05em{\sc i\kern-.025em b}\kern-.08em
    T\kern-.1667em\lower.7ex\hbox{E}\kern-.125emX}}
\begin{document}

\title{Online Object-Level Semantic Mapping for Quadrupeds in Real-World Environments}

\author{
\IEEEauthorblockN{Emad Razavi}
\IEEEauthorblockA{\textit{Dynamic Legged Systems}\\
\textit{Istituto Italiano di Tecnologia}}
\IEEEauthorblockA{\textit{DIBRIS, University of Genova}\\
Genoa, Italy\\
5782734@studenti.unige.it}
\and
\IEEEauthorblockN{Angelo Bratta}
\IEEEauthorblockA{\textit{Dynamic Legged Systems}\\
\textit{Istituto Italiano di Tecnologia}\\
Genoa, Italy\\
angelo.bratta@iit.it}
\and
\IEEEauthorblockN{João Carlos Virgolino Soares}
\IEEEauthorblockA{\textit{Dynamic Legged Systems}\\
\textit{Istituto Italiano di Tecnologia}\\
Genoa, Italy\\
joao.virgolino@iit.it} 
\and
\IEEEauthorblockN{\hspace*{11.0em} Carmine Recchiuto}
\IEEEauthorblockA{\hspace*{11.0em}\textit{RICE lab}\\
\hspace*{11.0em}\textit{University of Genova}\\
\hspace*{11.0em}Genoa, Italy\\
\hspace*{11.0em}carmine.recchiuto@unige.it}

\and
\IEEEauthorblockN{\hspace*{-4.0em} Claudio Semini}
\IEEEauthorblockA{\hspace*{-4.0em}\textit{Dynamic Legged Systems}\\
\hspace*{-4.0em}\textit{Istituto Italiano di Tecnologia}\\
\hspace*{-4.0em}Genoa, Italy\\
\hspace*{-4.0em}claudio.semini@iit.it}
}

\maketitle 

\begin{abstract}
We present an online semantic object mapping system for a quadruped robot operating in real indoor environments, turning sensor detections into named objects in a global map. During a run, the mapper integrates range geometry with camera detections, merges co-located detections within a frame, and associates repeated detections into persistent object instances across frames. Objects remain in the map when they are out of view, and repeated sightings update the same instance rather than creating duplicates. The output is a compact object layer that can be queried (class, pose, and confidence), is integrated with the occupancy map and readable by a planner. In on-robot tests, the layer remained stable across viewpoint changes. 

\end{abstract}
  

\section{Introduction}
Accurate mapping is fundamental for autonomous navigation. In complex environments featuring uneven terrain, low friction, gaps or stairs, wheeled robots struggle, while bipedal robots are usually unstable and slow. Quadrupeds, however, offer advantages in mobility and stability in such scenarios~\cite{Agha2022}.

In these settings, navigation goals are often tied to objects rather than geometric locations, e.g., “find the door”, “go to the toolbox”, or “stop by the charger.” A purely geometric map could be insufficient in such scenarios. Object-level semantics can help a planner decide where to go \cite{chaplot2020}. 

Earlier work showed that maps can combine geometry with object and location labels, giving planners a representation they can query directly for tasks like “go to the kitchen”\cite{kostavelis2016,crespo2017}. Learning-based methods train a policy that maintains a category-level semantic map from RGB–D (via both segmentation and detection) and conditioned on the goal class, scores frontiers or regions to select the next waypoint \cite{chaplot2020}. Multi-modal mapping fuses RGB-D images with LiDAR data to reduce depth measurement errors and can improve 3D placement. \cite{rollo2023}. On quadrupeds used in search-and-rescue scenarios, semantic layers focus on people and structural cues such as cracks, stairs, and doors, confirming the value of task-specific object information \cite{betta2024}. Recent open-vocabulary and vision–language methods add flexible labels and frontier scoring, but they often require comparatively heavy computation and dense 3D updates \cite{yokoyama2024,busch2024,qu2024}. In parallel, many practical Spot deployments such as industrial sites, tunnels, and labs, still rely on prior maps for stable localization in confined or hazardous spaces, which underlines the need for reliable, lightweight mapping substrates on real robots \cite{koval2022,Agha2022}. 

\begin{figure}[t]
  \centering
  \includegraphics[width=0.7\linewidth]{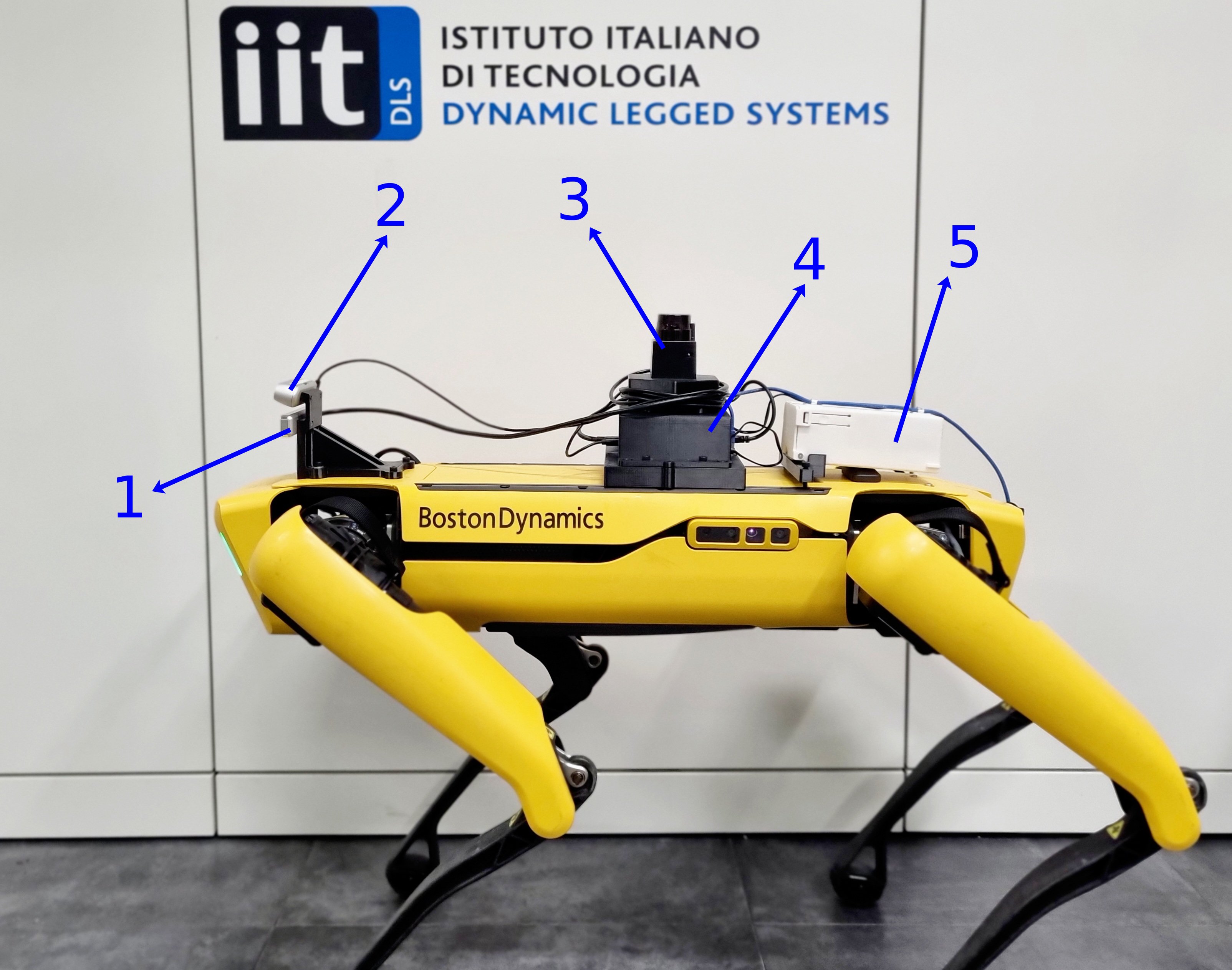}
    \caption{Spot by Boston Dynamics with onboard payload: 1. Intel RealSense T265 Camera, 2. Intel RealSense D435 Camera, 3. 2D LiDAR, 4. Intel NUC 11 i7-1165G7, and 5. A lithium-ion battery to power the NUC.}
  \label{fig:spot_deploy}
\end{figure}

\begin{figure*}[!t]
  \centering
  \includegraphics[width=0.85\linewidth]{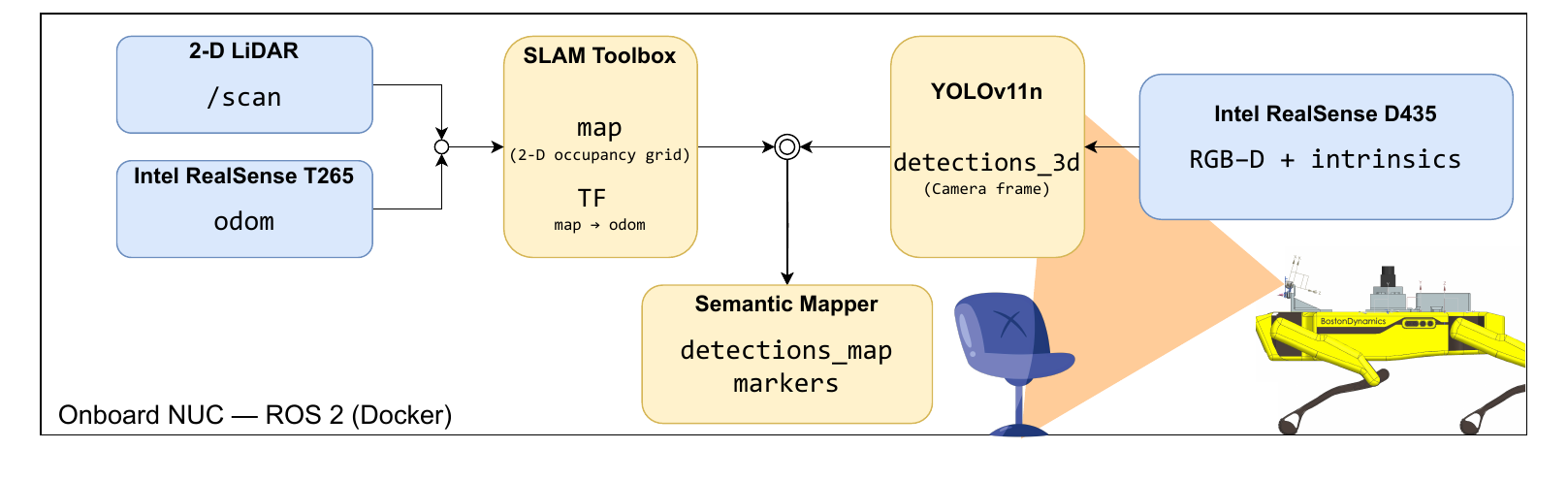}
  \caption{Overview of the proposed methodology. A 2D occupancy grid map is built using 2D LiDAR scans and odometry obtained from the T265 camera. Object detections are obtained from RGB images, and depth is used to estimate 3D positions. The semantic layer places detected objects on the map, merges nearby ones, and keeps them over time.}
  \label{fig:system_overview}
\end{figure*}

In this work, we present a lightweight semantic mapping method for a quadruped robot. The system fuses RGB-D camera object detections with LiDAR scans and T265 odometry to project the objects into a global map, merging nearby duplicates per frame, and associating repeated observations over time. Fig. \ref{fig:system_overview} summarizes the methodology. We avoid heavy 3D fusion and large vision–language scoring loops that raise latency and power draw \cite{yokoyama2024,busch2024,qu2024}, being consistent with onboard computational constraints \cite{koval2022,Agha2022}. The platform is a Boston Dynamics Spot equipped with a sensor payload and an onboard NUC (Fig.~\ref{fig:spot_deploy}). 

\section{Methods}
\subsection{Mapping (SLAM Toolbox with T265 odom and 2D LiDAR)}
We build a 2D occupancy map online with the \textit{SLAM Toolbox} \cite{Macenski2021}. A 2D LiDAR provides range geometry, while the Intel RealSense T265 provides visual–inertial odometry. The static transformations between the robot body frame, the LiDAR, and the T265 are fixed (calibrated once). During operation, the SLAM Toolbox uses visual odometry computed by the T265 tracking camera as a motion prior and refines it with scan matching and pose-graph loop closure, limiting drift and maintaining a globally consistent estimate. This occupancy map provides the geometric base for the semantic layer described next. 

\subsection{Visualization and Detection (D435 + YOLOv11)}
The Intel RealSense D435 provides synchronized RGB–D input for detection and projection: color images, depth aligned to the color stream, and the associated camera intrinsics~\cite{IntelD400Datasheet2023}. We also use a static transform between the robot body frame and the camera frame.

For object detection, we use  YOLOv11~\cite{Redmon2016}. Each color frame is preprocessed to the model input; outputs are class labels, scores, and 2D bounding boxes in the camera frame after non-maximum suppression. We publish a lightweight detection topic and an RViz overlay. Depth fusion and map anchoring are handled by the semantic layer in the next subsection.

\subsection{Semantic object layer: association and memory}
We transform each 3D detection into the map. We set the height (z-axis) to 0 and keep only the yaw rotation (roll and pitch are set to zero) to have a 2D map. We drop low-score detections by using per-class thresholds. Using Euclidean distance on the map, within the same frame, we remove near-duplicates by keeping only the highest-score detection for a class when two appear very close on the map. The key parameters used in our runs are summarized in Table~\ref{tab:core}.

\begin{center}
\captionsetup{type=table}
\caption{Core settings for association and memory.}
\label{tab:core}
\begin{tabular}{l r}
\toprule
Per-class confidence cutoff & 0.65 \\
Same-frame merge radius [m] & 0.20 \\
Reuse radius in map [m] & 0.80 \\
Promotion gate (hits / s / mean) & 10 / 2.0 / 0.50 \\
\bottomrule
\end{tabular}
\end{center}

\begin{figure}[H]
  \centering
  \includegraphics[width=0.9\columnwidth]{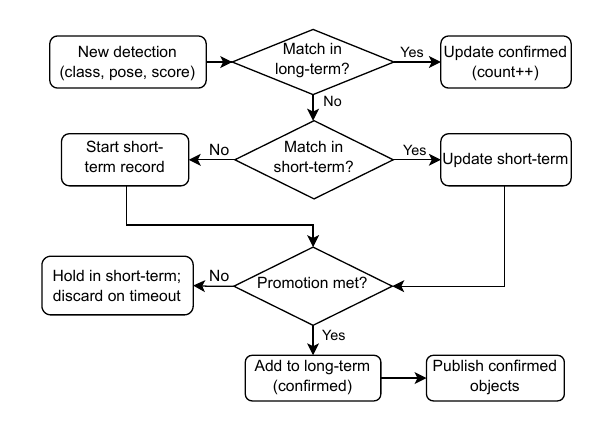}
  \caption{Association pipeline for confirmed objects.}
  \label{fig:assoc_flow}
\end{figure}

We have two memories: a short-term buffer of recent observations and a long-term list of confirmed objects. For each detection, we first check the long-term list with a nearest-neighbor search (NNS) in the 2D map. If a same-class object is nearby, we recognize it as seen again, increase its detection count by one, and leave its stored pose unchanged. Otherwise, we compare it with the short-term buffer: if no same class is within the radius, we start a new short-term record; if one is close, we update it with the new measurement. A new object is added to the long-term list only after the same class is re-observed several times within a short interval at approximately the same position and with sufficient confidence. Finally, we publish the current set of confirmed objects in the map frame. Fig. \ref{fig:assoc_flow} summarizes the association and memory logic.

\FloatBarrier

\section{Results}
\begin{figure}[t]
  \centering
  \subfloat[RViz—map with confirmed objects]{%
    \includegraphics[width=0.49\linewidth]{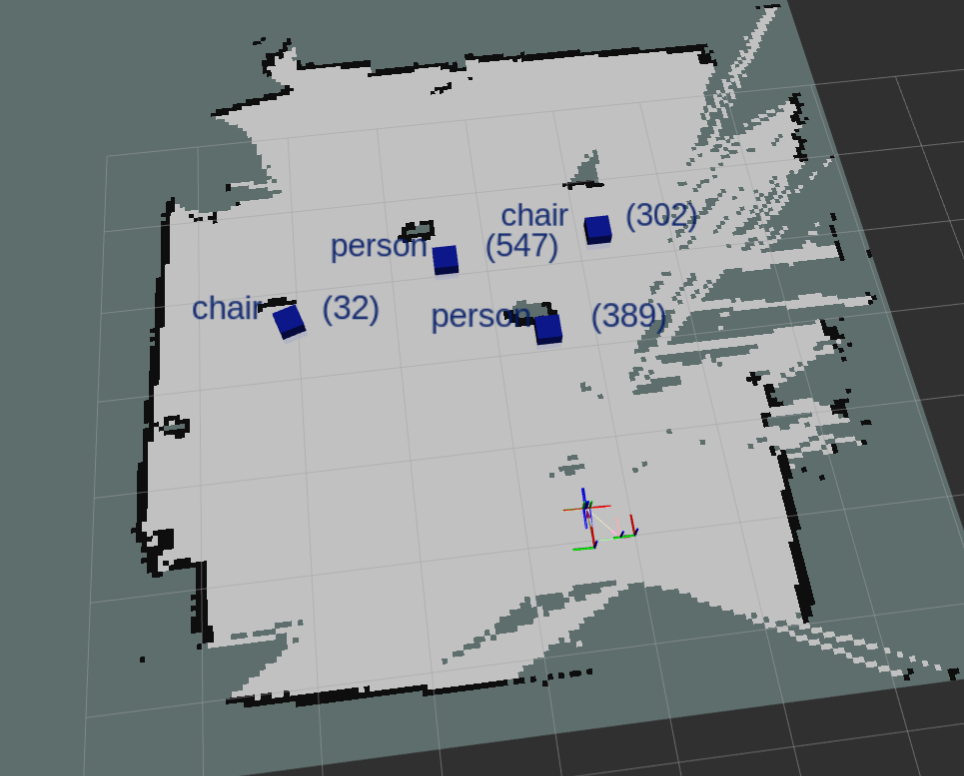}%
  }\hfill
  \subfloat[Lab scene (people, chairs)]{%
    \includegraphics[width=0.51\linewidth]{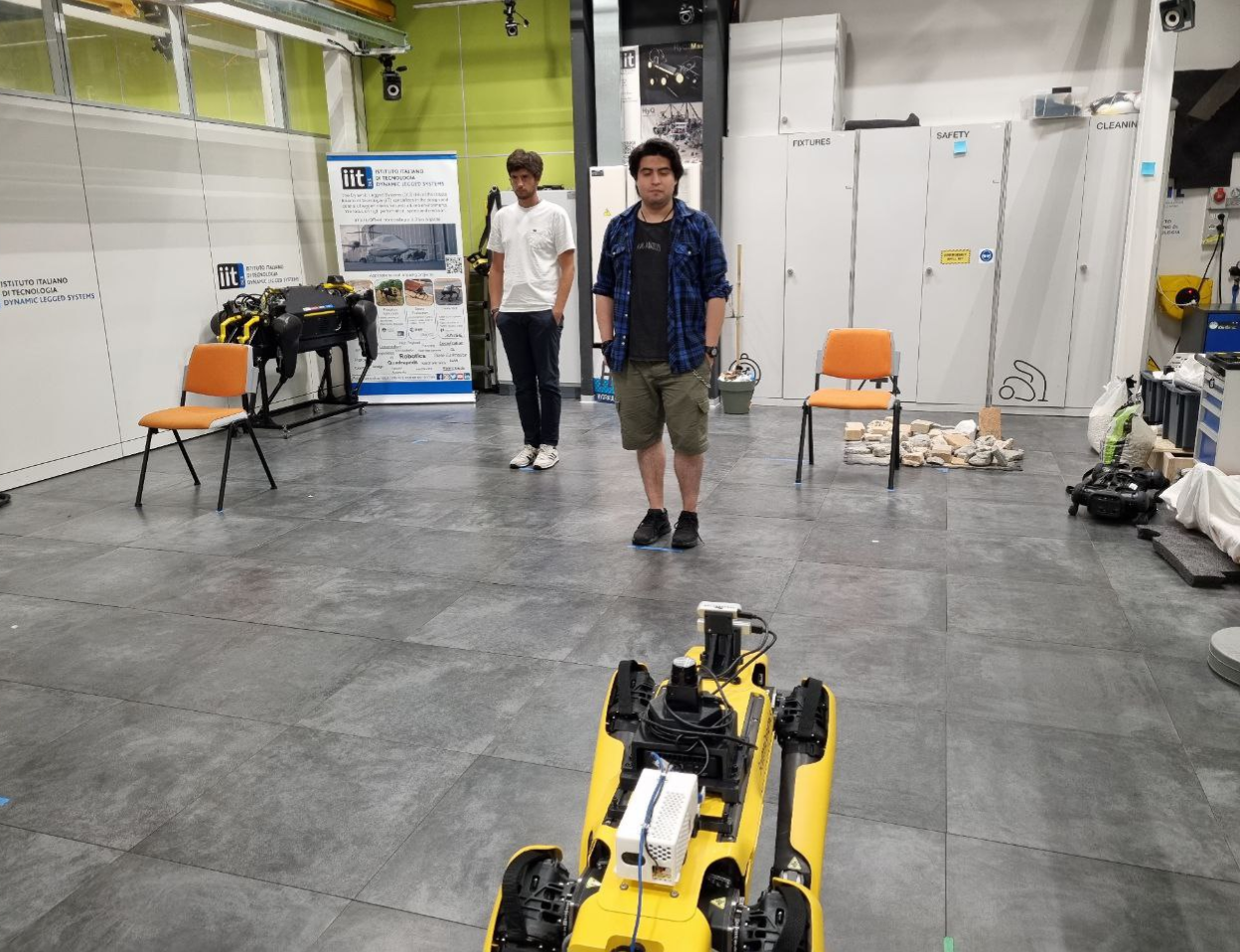}%
  }
  \caption{Setup and output. The semantic layer tracked only \texttt{person} and \texttt{chair}.}
  \label{fig:results}
\end{figure}

\begin{figure}[t]
  \centering
  \includegraphics[width=0.9\linewidth]{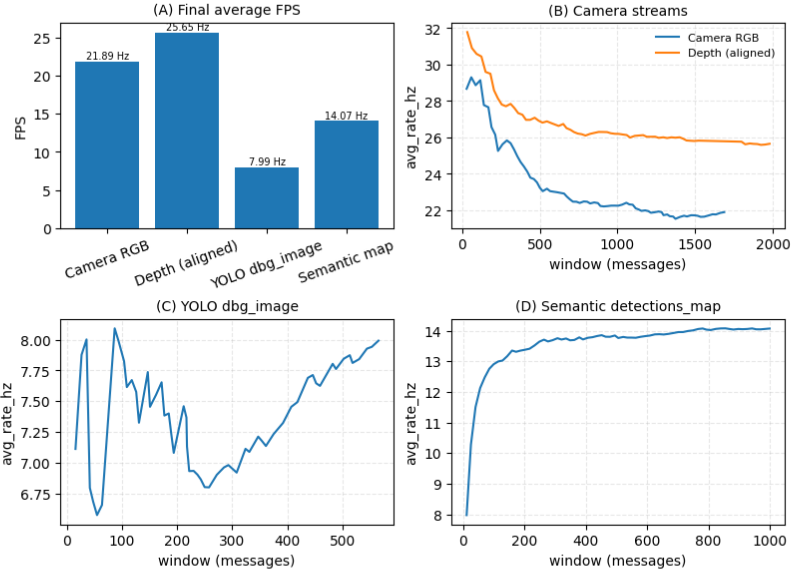}
    \caption{(A) Final FPS. (B) Camera RGB and aligned depth, average rate  vs.\ cumulative messages (window). (C) YOLO debug-image rate. (D) Semantic object-map rate. Note: the YOLO stream publishes only when detections exist.}  
    \label{fig:time_frames}
\end{figure}

We mounted a D435 RGB–D, a T265, and a 2D LiDAR on the quadruped and ran the SLAM Toolbox with our semantic mapper inside a ROS~2 container. An operator drove the robot with a joystick through a lab with two people and two chairs, as shown in Fig. \ref{fig:results}(b). As the robot moved, the SLAM system built a 2D occupancy map, while the semantic mapper projected detections into the map in real time\footnote{Demo video: \href{https://youtu.be/F4q-SN-NFhI}{online}.}. Using the depth images, we computed a 3D position for each detection and transformed it into the global map via the calibrated camera–to–body transform and the SLAM-estimated robot pose. In RViz, blue cubes, shown in Fig. \ref{fig:results}(a), mark confirmed objects. Each one is labeled with its class and hit count. Repeated sightings update the same instance, and per-frame near-duplicate boxes are suppressed to the highest-score to keep the map uncluttered. Fig. \ref{fig:time_frames} shows the asynchronous frame rates of the visualization topics. The semantic layer maintains stable instances despite these mismatched stream rates.

\section{Conclusion}
We built an online, object-level semantic map for a quadruped. Geometry comes from the SLAM Toolbox (2D occupancy with T265 and a 2D LiDAR) and detections from a D435 with YOLO are anchored in the map frame as persistent instances. Simple map-frame association plus short- and long-term memory keeps objects stable through viewpoint changes while avoiding per-frame duplicates. An autonomous navigation system that uses this layer is left for future work.

The limitation we observed is depth misalignment: D435 depth is not consistent with the LiDAR scan plane, which introduces range/pose bias when placing objects in the map. Other limitations are reliance on detector centers, purely geometric association (no appearance cues), and evaluation limited to indoor runs.

\section{Acknowledgment}
We thank Gabriel Fischer Abati and Miguel Fernandes for their valuable assistance during the experiments.

This work was also carried out within the framework of the project ``RAISE - Robotics
and AI for Socio-economic Empowerment” and has been supported by European Union – NextGenerationEU.

\vspace{12pt}
\end{document}